\documentclass[letterpaper]{article} 
\usepackage{aaai24}  
\usepackage{times}  
\usepackage{helvet}  
\usepackage{courier}  
\usepackage[hyphens]{url}  
\usepackage{graphicx} 
\usepackage{natbib}  
\usepackage{caption} 
\usepackage{algorithm}
\usepackage{algorithmic}
\usepackage{amsmath,amsfonts}
\usepackage{multirow}
\usepackage{tabularx}
\usepackage{booktabs}
\usepackage{colortbl}
\usepackage[table]{xcolor}
\usepackage{makecell}
\usepackage{newfloat}
\usepackage{listings}
\DeclareCaptionStyle{ruled}{labelfont=normalfont,labelsep=colon,strut=off} 
\floatstyle{ruled}
\newfloat{listing}{tb}{lst}{}
\floatname{listing}{Listing}
\title{AVSegFormer: Audio-Visual Segmentation with Transformer}
\author{
    Shengyi Gao$^1$,
    Zhe Chen$^1$,
    Guo Chen$^1$,
    Wenhai Wang$^2$,
    Tong Lu$^{1}$\thanks{corresponding author}
}
\affiliations{
    \textsuperscript{\rm 1}State Key Lab for Novel Software Technology, Nanjing University\\
    \textsuperscript{\rm 2}The Chinese University of Hong Kong\\

}

\usepackage{bibentry}

\definecolor{baselinecolor}{gray}{.9}
\usepackage{subfig}

\def\eg{\emph{e.g.}}  

\def\ie{\emph{i.e.}}

\begin{document}

\maketitle

\begin{abstract}
Audio-visual segmentation (AVS) aims to locate and segment the sounding objects in a given video, which demands audio-driven pixel-level scene understanding. The existing methods cannot fully process the fine-grained correlations between audio and visual cues across various situations dynamically. They also face challenges in adapting to complex scenarios, such as evolving audio, the coexistence of multiple objects, and more. In this paper, we propose AVSegFormer, a novel framework for AVS that leverages the transformer architecture. Specifically, It comprises a dense audio-visual mixer, which can dynamically adjust interested visual features, and a sparse audio-visual decoder, which implicitly separates audio sources and automatically matches optimal visual features. Combining both components provides a more robust bidirectional conditional multi-modal representation, improving the segmentation performance in different scenarios.
Extensive experiments demonstrate that AVSegFormer achieves state-of-the-art results on the AVS benchmark.
The code is available at \url{https://github.com/vvvb-github/AVSegFormer}.
\end{abstract}

\section{Introduction}

Just as humans effortlessly establish meaningful connections between audio and visual signals, capturing the rich information they convey, the intertwined modalities of audio and vision play pivotal roles in observing and comprehending the real world.
Based on this insight, a wide range of audio-visual understanding tasks, such as audio-visual correspondence \cite{arandjelovic2017look,arandjelovic2018objects}, audio-visual event localization \cite{lin2019dual,lin2020audiovisual}, audio-visual video parsing \cite{tian2020unified,wu2021exploring}, and sound source localization \cite{arandjelovic2017look,arandjelovic2018objects} have been proposed and actively explored in recent research.

\begin{figure}[t]
  \centering
  \includegraphics[width=0.95\linewidth]{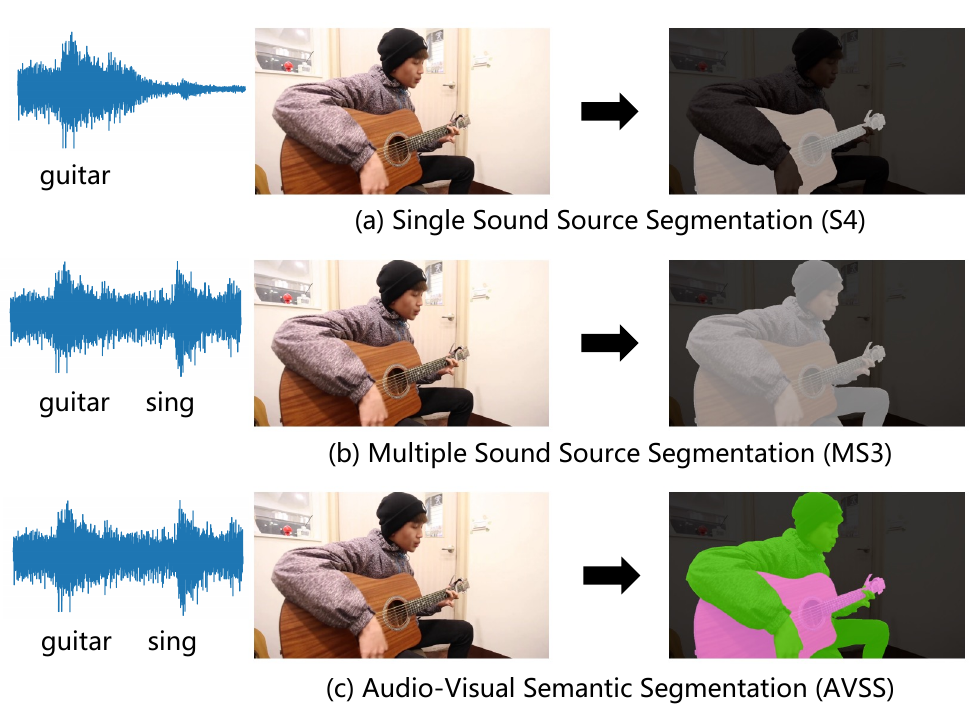}
  \caption{Illustration of audio-visual segmentation (AVS).
  AVS aims to segment sounding objects from video frames according to the given audio.
  In the S4 sub-task, the input audio only contains one sound source, while in MS3 the input audio has multiple sound sources. 
    Besides, S4 and MS3 only require binary segmentation, whereas AVSS requires more difficult multiple-category semantic segmentation.
  }
  \label{fig:dataset}
\end{figure}

Unlike these coarse-grained tasks, audio-visual segmentation (AVS) \cite{zhou2023audio} proposes more fine-grained perceptive goals, aiming to locate the audible frames and delineate the shape of the sounding objects \cite{zhou2022audio,zhou2023audio}.
To be more specific, this task involves three sub-tasks: single sound source segmentation (S4), multiple sound source segmentation (MS3), and audio-visual semantic segmentation (AVSS).
Figure~\ref{fig:dataset} illustrates the objectives of the three sub-tasks. Their fine-grained perceptual goal demands the model to possess the capability to discern the intricate relationship between each image pixel and audio information. However, the existing methods \cite{chen2021localizing, qian2020multiple, mahadevan2020making} developed for other audio-visual tasks face challenges when directly applied in this context.

\begin{figure*}[h]
  \centering
  \includegraphics[width=0.99\linewidth]{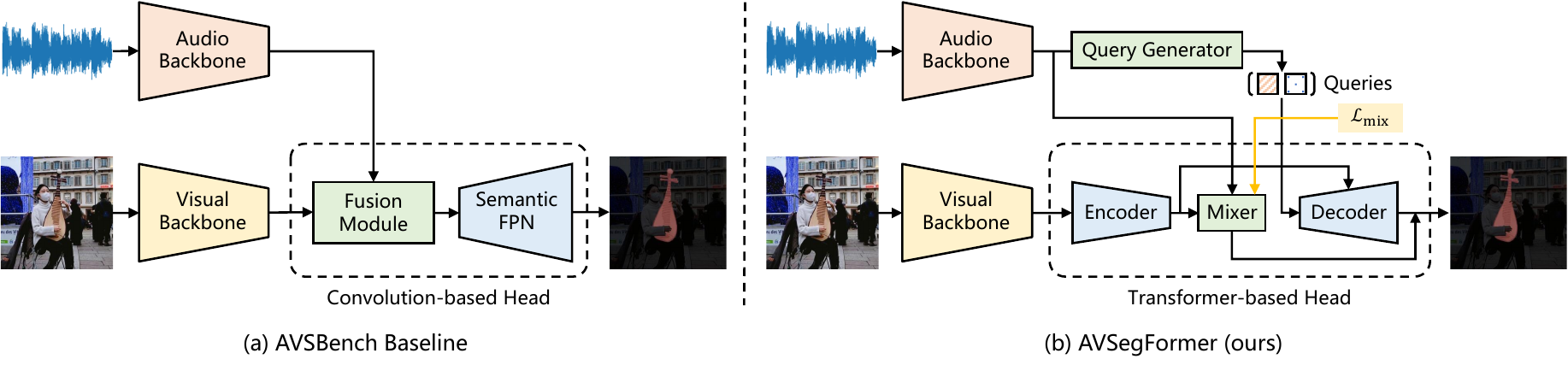}
  \caption{
  Overview of the AVSBench baseline and our AVSegFormer.
  (a) The baseline method \cite{zhou2022audio} incorporates a modality fusion module before Semantic FPN \cite{li2022panoptic}. (b) The proposed AVSegFormer performs audio-visual segmentation with transformer-based architecture. It has four key designs to significantly improve performance, including (1) a transformer encoder, (2) an audio-visual mixer, (3) a query generator, and (4) a cross-modal transformer decoder.}
  \label{fig:comp}
\end{figure*}

AVSBench \cite{zhou2023audio} first proposes the fine-grained perceptive method for AVS tasks that achieves state-of-the-art audio-visual segmentation performance. 
Figure~\ref{fig:comp}(a) illustrates its network architecture, which incorporates a modality fusion module before Semantic FPN \cite{kirillov2019panoptic} to enable audio-visual segmentation.
This method is simple and effective yet falls short of fully mining the fine-grained correlations between audio and visual cues across various situations.
First, a series of targets sound simultaneously in scenes of multiple sound sources. 
The mixed audio signal with higher information density is difficult to attend to the visual signal adaptively.
Second, the simple-fusion model may make it difficult to extract audio information bound with the corresponding frame when the sound source objects change over time. 
Additionally, in cases where multiple objects coexist within a single contextual frame (\eg, two guitars or one person and one guitar), the vanilla dense segmentation method struggles to untangle the different sound sources in the audio, making it challenging to achieve precise localization and segmentation in a one-to-one manner.

To remedy these issues, we propose \textbf{AVSegFormer}, a novel framework for audio-visual segmentation with the transformer architecture. The brief architecture is shown in Figure~\ref{fig:comp}(b). 
AVSegFormer comprises four key components: (1) a transformer encoder building the mask feature, (2) an audio-visual mixer generating the vision-conditioned mask feature, (3) a query generator initializing sparse audio-conditioned queries, and (4) a cross-modal transformer decoder separating potential sparse object in the visual feature. Among them, we design an auxiliary mixing loss to supervise the cross-modal feature generated by the audio-visual mixer. It encourages the model to attend to useful information within complex audio semantics and predict the segmentation mask densely. Compared to the dense vision-conditioned mixers, the cross-modal transformer decoder aims to build potential sparse queries with the reversed condition. It implicitly separates audio sources and automatically matches optimal visual features for different queries.
These queries will be combined with the vision-conditioned feature map through the matrix production. 
At last, the fused bidirectional conditional feature maps will be used for the final segmentation.

Overall, our contributions to this work are three-fold:

(1) We propose AVSegFormer, a transformer-based method for three scenarios of the audio-visual segmentation tasks. It combines bidirectional conditional cross-modal feature fusion to provide a more robust audio-visual segmentation representation.

(2) We propose a dense audio-visual mixer and a sparse audio-visual decoder to provide pixel-level and instance-level complementary representations that can efficiently adapt the scenarios of multiple sound sources and objects.

(3) Extensive experiments on three sub-tasks of AVS are conducted, demonstrating that AVSegFormer significantly outperforms existing state-of-the-art methods \cite{mao2023contrastive, zhou2022audio}. 

\begin{figure*}[h]
  \centering
  \includegraphics[width=0.9\linewidth]{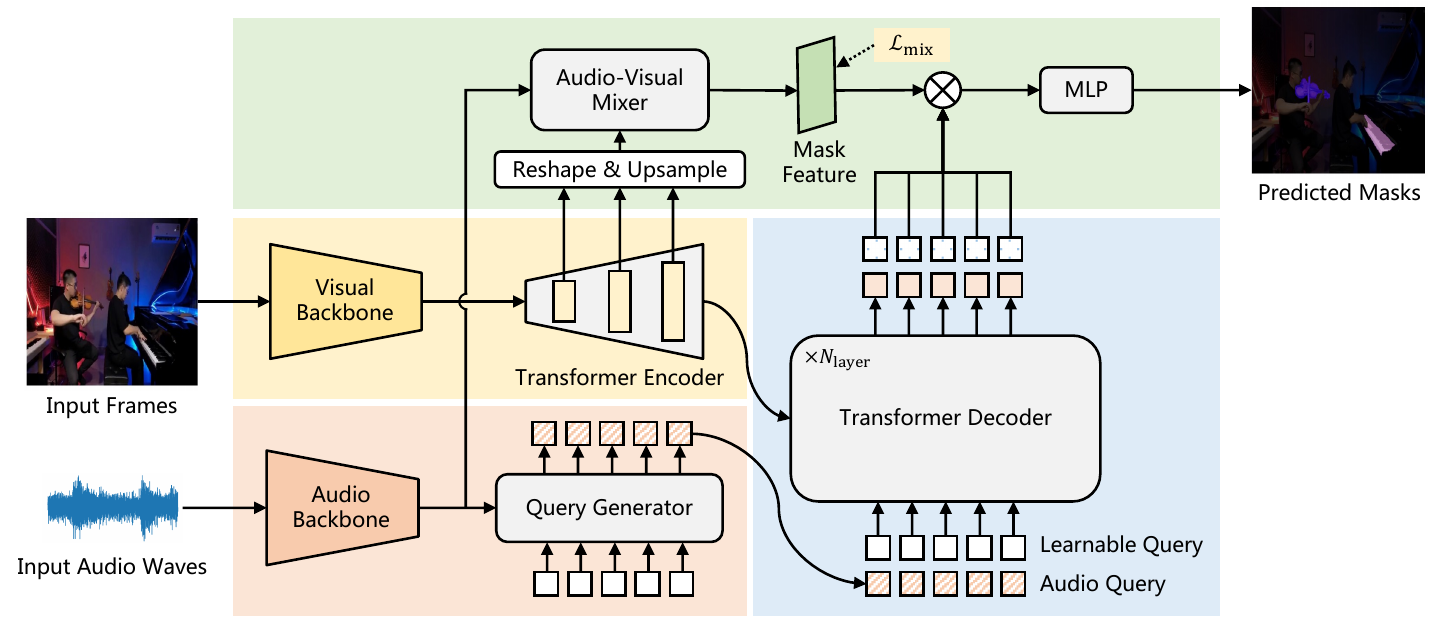}
  \caption{Overall architecture of AVSegFormer. We propose four key components in this framework: 
  (1) The transformer encoder builds the initial mask feature;
  (2) An audio-visual mixer with an auxiliary mixing loss $\mathcal{L}_{\rm mix}$ generates the vision-conditioned mask feature.
  (3) The query generator initializes sparse audio-conditioned queries, enabling the model to recognize abundant auditory semantics. 
  (4) The cross-modal transformer decoder separates potential sparse objects in the visual feature.}
  \label{fig:arch}
\end{figure*}

\section{Related Works}

\vspace{0.5em}
\subsection{Multi-Modal Tasks}
In recent years, multi-modal tasks have gained significant attention in the research community.
Among these, text-visual tasks have attracted considerable interest from researchers. 
Numerous works focus on related tasks, such as visual question answering \cite{antol2015vqa, wu2016ask} and visual grounding \cite{deng2021transvg, kamath2021mdetr}.
In addition to text-visual tasks, audio-visual tasks are emerging as hot spots. 
Related tasks include audio-visual correspondence \cite{arandjelovic2017look,arandjelovic2018objects}, audio-visual event localization \cite{lin2019dual,lin2020audiovisual}, and sound source localization \cite{arandjelovic2017look,arandjelovic2018objects}. 
Concurrently, many works \cite{zhu2022uni, wang2022ofa} have proposed unified architectures to deal with multi-modal inputs.

Most of these works are based on transformer architecture \cite{vaswani2017attention}, demonstrating a strong cross-modal capability. 
Their success highlights the reliability of transformers in the multi-modal field. 
As a recently proposed multi-modal task, audio-visual segmentation \cite{zhou2022audio,zhou2023audio} shares many commonalities with the aforementioned tasks. The pioneering works in these areas have significantly inspired our research of AVSegFormer.

\subsection{Vision Transformer}
During the past few years, Transformer \cite{vaswani2017attention} has experienced rapid development in natural language processing. 
Following this success, the Vision Transformer (ViT) \cite{dosovitskiy2020image} emerged, bringing the transformer into the realm of computer vision and yielding impressive results. 
Numerous works \cite{liu2021swin,wang2022pvt,chen2022vision} have built upon ViT, leading to the maturation of vision transformers, especially in object detection and image segmentation tasks.
As the performance of vision transformers continues to advance, they are increasingly replacing CNNs as the mainstream paradigm in the field of computer vision, especially in object detection and image segmentation tasks.

For downstream tasks, \citet{carion2020end} proposed the DETR model and designed a novel bipartite matching loss based on the transformer architecture.
Subsequently, improved frameworks such as Deformable DETR \cite{zhu2020deformable} and DINO \cite{zhang2022dino} are proposed, introducing mechanisms like deformable attention and denoise training. 
These arts take vision transformers to new heights. 
The remarkable performance of vision transformers has also inspired us to apply this paradigm to AVS tasks, anticipating further advancements in the field.

\subsection{Image Segmentation}
Image segmentation is a critical visual task that involves partitioning an image into distinct segments or regions. 
It includes three different tasks: instance segmentation, semantic segmentation, and panoptic segmentation. 
Early research proposed specialized models for these tasks, such as Mask R-CNN \cite{he2017mask} and HTC \cite{chen2019hybrid} for instance segmentation, or FCN \cite{long2015fully} and U-Net \cite{ronneberger2015u} for semantic segmentation. 
After panoptic segmentation was proposed, some related research \cite{kirillov2019panoptic,xiong2019upsnet,li2022panoptic} were conducted and designed universal models for both tasks. 

The recent introduction of the transformer has led to the development of new models that can unify all the segmentation tasks. 
Mask2Former \cite{cheng2022masked} is one such model that introduces mask attention into the transformer.
Mask DINO \cite{li2022mask} is a unified transformer-based framework for both detection and segmentation.
Recently, OneFormer \cite{jain2022oneformer} presented a new universal image segmentation framework with transformers.
These models have brought image segmentation to a new level.
Considering that the AVS task involves segmentation, these methods have significantly contributed to our work.

\section{Methods}

\subsection{Overall Architecture}
Figure~\ref{fig:arch} illustrates the overall architecture of our method. In contrast to previous CNN-based methods \cite{zhou2022audio, zhou2023audio}, we design a query-based framework to leverage the transformer architecture. 
Specifically, the query generator initializes audio queries, and the transformer encoder extracts multi-scale features, which serve as inputs of the transformer decoder for separating potential sparse objects. 
Besides, the audio-visual mixer will further amplify relevant features and suppress irrelevant ones, while the auxiliary mixing loss helps supervise the reinforced features.

\subsection{Multi-Modal Representation}

\textbf{Visual encoder.}
We follow the feature extraction process adopted in previous methods \cite{zhou2022audio,zhou2023audio}, which uses a visual backbone and an audio backbone to extract video and audio features, respectively. 
The dataset provides pre-extracted frame images from videos, making the process similar to image feature extraction. 
Specifically, the input video frames are denoted as $x_{\rm visual}\in \mathbb{R}^{T\times3\times H\times W}$, in which $T$ denotes the number of frames. 
Then, we use a visual backbone (\eg, ResNet-50 \cite{he2016deep}) to extract hierarchical visual features $\mathcal{F}_{\rm visual}$, which can be written as:
\begin{equation}
    \mathcal{F}_{\rm visual}=\{\mathcal{F}_1,\mathcal{F}_2,\mathcal{F}_3,\mathcal{F}_4\}, 
\end{equation}
in which $\mathcal{F}_i\in \mathbb{R}^{T\times 256\times \frac{H}{2^{i+1}}\times \frac{W}{2^{i+1}}}$ and $i \in [1, 2, 3, 4]$.

\textbf{Audio encoder.}
The process of audio feature extraction follows the VGGish \cite{hershey2017cnn} method. Initially, the audio is resampled to 16kHz mono audio $x_{\rm audio}\in \mathbb{R}^{N_{\rm samples}\times 96\times 64}$, where $N_{\rm samples}$ is related to the audio duration. 
Then, we perform a short-time Fourier transform to obtain a mel spectrum. 
The mel spectrum is calculated by mapping the spectrum to a 64th-order mel filter bank and then fed into the VGGish model to obtain the audio features $\mathcal{F}_{\rm audio}\in \mathbb{R}^{T\times 256}$, where $T$ means the number of frames.

\begin{figure}[t]
  \centering
  \vspace{0.5em}
  \includegraphics[width=1\linewidth]{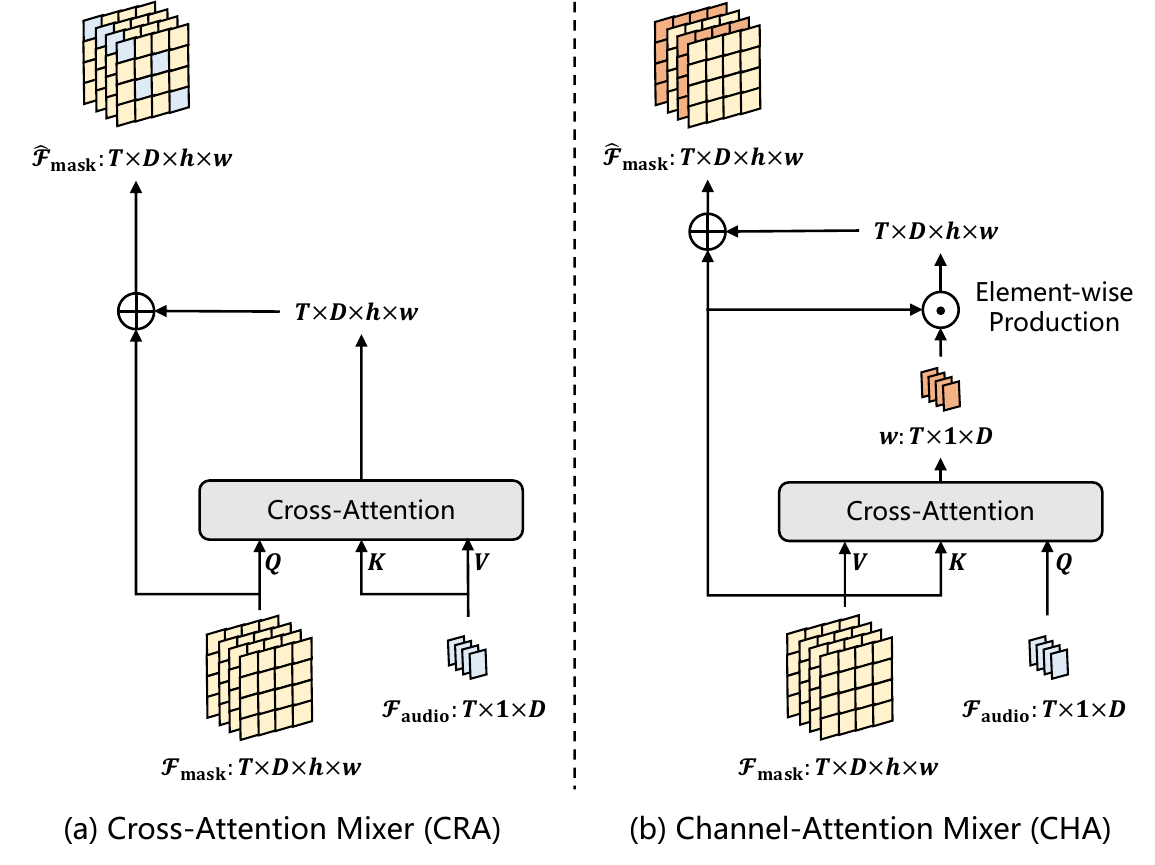}
  \caption{Architecture of the audio-visual mixer. (a) Our initial design incorporates a cross-attention mixer, which fails to deliver satisfactory results. (b) We ultimately adopted the channel-attention mixer design, demonstrating significantly improved performance.}
  \label{fig:fusion}
\end{figure}

\subsection{Query Generator}
The query generator is designed to generate sparse audio-conditioned queries, which helps the model fully understand the auditory information. 
At the beginning, we have an initial query $Q_{\rm init}\in \mathbb{R}^{T\times N_{\rm query}\times D}$ and audio feature $\mathcal{F}_{\rm audio}\in \mathbb{R}^{T\times D}$, here $N_{\rm query}$ represents the number of queries. We employ $Q_{\rm init}$ as queries and $\mathcal{F}_{\rm audio}$ as keys and values, feed them into the query generator, and obtain the audio query $Q_{\rm audio}\in\mathbb{R}^{T\times N_{\rm query}\times D}$. Finally, we incorporate audio query $Q_{\rm audio}$ and learnable query $Q_{\rm learn}$ as a mixed query $Q_{\rm mixed}$ for the input of the transformer decoder. 

The addition of learnable queries enhances our model's adaptability for various AVS tasks and datasets. It enables the model to learn dataset-level contextual information, and adjust the attention allocated to different sounding targets.

\subsection{Transformer Encoder}
The transformer encoder is responsible for building the mask feature. 
Specifically, we collect the backbone features of three resolutions (\ie, 1/8, 1/16, and 1/32), and then flatten and concatenate them as the input queries for the transformer encoder. 
After that, the output features are reshaped to their original shapes, and the 1/8-scale features are taken out separately and 2$\times$ upsampled.
Finally, we add the upsampled features to the 1/4-scale features from the visual backbone and obtain the mask feature $\mathcal{F_{\rm mask}}\in \mathbb{R}^{T\times D\times h\times w}$, where $h=\frac{H}{4},w=\frac{W}{4}$, and $D$ is the embed dimension. 

\subsection{Audio-Visual Mixer}

As illustrated in Figure~\ref{fig:arch}, the segmentation mask is generated based on the mask feature, which plays a crucial role in the final results. 
However, since the audio semantics can vary widely, a static network may not be able to capture all of the relevant information.
This limitation may hinder the model's ability to identify inconspicuous sounding objects.

To address this issue, we propose an audio-visual mixer as shown in Figure~\ref{fig:fusion}(b). 
The design of this module is based on channel attention, which allows the model to selectively amplify or suppress different visual channels depending on the audio feature, improving its ability to capture complex audio-visual relationships.
Specifically, the mixer learns a set of weights $\omega$ through audio-visual cross-attention, and applies them to highlight the relevant channels. 
The whole process can be represented as follows:
\begin{equation}
    \begin{split}
        & \omega = \mathrm{softmax}(\frac{\mathcal{F_{\rm audio}}\mathcal{F_{\rm mask}}^T}{\sqrt{D/n_{\rm head}}})\mathcal{F_{\rm mask}}, \\
        & \mathcal{\hat{F}_{\rm mask}}=\mathcal{F_{\rm mask}}+\mathcal{F_{\rm mask}}\odot \omega.
    \end{split}
\end{equation}
Here, $\mathcal{F}_{\rm audio}$ and $\mathcal{F}_{\rm mask}$ represent the input audio feature and the initial mask feature, and $\mathcal{\hat{F}}_{\rm mask}$ denotes the mixed mask feature.
$n_{\rm head}$ means the number of attention heads, which is set to 8 by default following common practice.

\subsection{Transformer Decoder}
The transformer decoder is designed to build potential sparse queries, and optimally match the visual features with corresponding queries.
We utilize the mixed query $Q_{\rm mixed}$ as the input query and the multi-scale visual features as key/value. 
As the decoding process continues, the output queries $Q_{\rm output}$ continuously aggregate with visual features, ultimately combining the auditory and visual modalities and containing various target information.

To generate the segmentation masks, we multiply the mask feature $\mathcal{\hat{F}_{\rm mask}}\in \mathbb{R}^{T\times D\times h\times w}$ obtained from the audio-visual mixer with the mixed queries $Q_{\rm output}$ from the query generator. 
Then, an MLP is used to integrate different channels. 
Finally, the model predicts the mask $\mathcal{M}$ through a fully connected layer:
\begin{equation}
    \mathcal{M}={\rm FC}(\mathcal{\hat{F}_{\rm mask}}+{\rm MLP}(\mathcal{\hat{F}_{\rm mask}}\cdot Q_{\rm output})).
\end{equation}
Here, $\rm MLP(\cdot)$ represents the MLP layer, and $\rm FC(\cdot)$ means the fully connected layer. The output $\mathcal{M}\in \mathbb{R}^{T\times N_{\rm class}\times h\times w}$ is the predicted segmentation mask, with the dimension $N_{\rm class}$ denotes the number of semantic classes.

\subsection{Loss Function}

\textbf{Auxiliary mixing loss.}
With the introduction of the audio-visual mixer, our model has maintained great capability in dealing with plentiful audio semantics, but its robustness is still insufficient when facing complex scenes. Thus, we design a mixing loss $\mathcal{L}_{\rm mix}$ to supervise the mixer, enabling the model to more accurately locate target objects. Specifically, we integrate all channels of the mask feature $\mathcal{\hat{F}_{\rm mask}}$ through a linear layer and predict a binary mask. At the same time, we extract all foreground labels in the ground truth as a new binary label and calculate the Dice loss \cite{milletari2016vnet} between them.

\textbf{Total loss.}
The loss function comprises two parts: IoU loss and mixing loss. 
The IoU loss $\mathcal{L}_{\rm IoU}$ is calculated by comparing the final segmentation mask with the ground truth. Here, we use Dice loss \cite{milletari2016vnet} for supervision. Considering that in AVS tasks, the proportion of segmented objects occupying the entire image is relatively small, the model can better focus on the foreground and reduce interference from the background by using Dice loss.
Thus, the total loss of our method is:
\begin{equation}
    \mathcal{L}=\mathcal{L}_{\rm IoU}+\lambda\mathcal{L}_{\mathrm{mix}}.
\end{equation}
Here, $\lambda$ is a coefficient that controls the effect of the auxiliary loss. We set $\lambda=0.1$ as it performs best.

\subsection{Discussion}

AVSegFormer adopts a framework for segmentation that resembles Mask2Former \cite{cheng2022masked}. However, it distinguishes itself by tailoring enhancements specifically for AVS tasks, accommodating the input of multi-modal information, which is a capability not inherent in Mask2Former. We introduce dual-tower backbone networks to ensure comprehensive extraction of both visual and auditory features. Besides, we devise a novel dense audio-visual mixer and a sparse audio-visual decoder that empower the proposed AVSegFormer to leverage auditory cues effectively, resulting in enhanced segmentation performance. These tailored designs hold substantial implications for addressing the emerging challenges of the AVS tasks.

\section{Experiments}

\begin{table*}[t]\small
  \centering
  \vspace{0.4em}
  \setlength{\tabcolsep}{2.6mm}
  \begin{tabular}{ll|cc|cc|cc|c}
    \toprule
     \multirow{2}{*}{Method} & \multirow{2}{*}{Backbone} & \multicolumn{2}{c|}{S4} & \multicolumn{2}{c|}{MS3} & \multicolumn{2}{c|}{AVSS} & \multirow{2}{*}{Reference}\\
     & & F-score & mIoU & F-score & mIoU & F-score & mIoU & \\
    \midrule 
     LVS  & ResNet-50  & 51.0 & 37.94 & 33.0 & 29.45 & $-$ & $-$ & CVPR'2021 \\
     MSSL  & ResNet-18  & 66.3 & 44.89 & 36.3 & 26.13 & $-$ & $-$ & ECCV'2020 \\
     3DC  & ResNet-34  & 75.9 & 57.10 & 50.3 & 36.92 & 21.6 & 17.27 & BMVC'2020 \\
     SST  & ResNet-101  & 80.1 & 66.29 & 57.2 & 42.57 & $-$ & $-$ & CVPR'2021 \\
     AOT  & Swin-B  & $-$ & $-$ & $-$ & $-$ & 31.0 & 25.40 & NeurIPS'2021 \\
     iGAN  & Swin-T  & 77.8 & 61.59 & 54.4 & 42.89 & $-$ & $-$ & ArXiv'2022 \\
     LGVT  & Swin-T  & 87.3 & 74.94 & 59.3 & 40.71 & $-$ & $-$ & NeurIPS'2021 \\
    \midrule
     AVSBench-R50  & ResNet-50  & 84.8 & 72.79 & 57.8 & 47.88 & 25.2 & 20.18 & ECCV'2022 \\
     DiffusionAVS-R50 & ResNet-50 & \textbf{86.9} & 75.80 & 62.1 & 49.77 & $-$ & $-$ & ArXiv'2023 \\
     \rowcolor{gray!20}
     AVSegFormer-R50 (ours) & ResNet-50  & 85.9 & \textbf{76.45} & 62.8 & 49.53 & 29.3 & 24.93 & AAAI'2024 \\
     \rowcolor{gray!20}
     AVSegFormer-R50* (ours) & ResNet-50  & 86.7 & 76.38 & \textbf{65.6} & \textbf{53.81} & \textbf{31.5} & \textbf{26.58} & AAAI'2024 \\
     \midrule
     AVSBench-PVTv2  & PVTv2 & 87.9 & 78.74 & 64.5 & 54.00 & 35.2 & 29.77 & ECCV'2022 \\
     DiffusionAVS-PVTv2 & PVTv2 & 90.2 & 81.38 & 70.9 & 58.18 & $-$ & $-$ & ArXiv'2023\\
     \rowcolor{gray!20}
     AVSegFormer-PVTv2 (ours) & PVTv2 & 89.9 & 82.06 & 69.3 & 58.36 & 42.0 & 36.66 & AAAI'2024 \\
     \rowcolor{gray!20}
     AVSegFormer-PVTv2* (ours) & PVTv2 & \textbf{90.5} & \textbf{83.06} & \textbf{73.0} & \textbf{61.33} & \textbf{42.8} & \textbf{37.31} & AAAI'2024 \\
  \bottomrule
\end{tabular}
  \caption{Comparison with state-of-the-art methods on the AVS benchmark.
  All methods are evaluated on three AVS sub-tasks, including single sound source segmentation (S4), multiple sound source segmentation (MS3), and audio-visual semantic segmentation (AVSS).
  The evaluation metrics are F-score and mIoU. The higher the better.
  *We tried to enlarge the image resolution to 512$\times$512. 
  }
  \label{tab:comp}
\end{table*}

\begin{table*}[t]\small
\vspace{1em}
\centering
\subfloat[
    Effect of the number of queries. We find that 300 queries work better than other settings.
    \label{tab:abla1}
]{
    \centering
    \begin{minipage}[t]{0.22\linewidth}{
    \vspace{0pt}
    \begin{center}
    \setlength{\tabcolsep}{0.3mm}
    \begin{tabular}{c|cc|cc}
    \toprule
     \multirow{2}{*}{$N_{\rm query}$}  & \multicolumn{2}{c|}{S4}  & \multicolumn{2}{c}{MS3}   \\
       & mIoU & F & mIoU & F \\
    \midrule
     1   & 79.6 & 86.6 & 59.1 & 69.7 \\
     100 & 81.4 & 88.9 & 60.5 & 71.2 \\
     200 & 82.3 & 89.6 & 61.0 & 72.4 \\
     \rowcolor{gray!20}
     300 & 83.1 & 90.5 & 61.3 & 73.0 \\
    \bottomrule
    \end{tabular}
    \end{center}}
    \end{minipage}
}
\hspace{1em}
\subfloat[
    Effect of the learnable query. Using learnable queries along with audio queries improves the performance.
    \label{tab:abla2}
]{
    \centering
    \begin{minipage}[t]{0.23\linewidth}{
    \vspace{0pt}
    \setlength{\tabcolsep}{0.3mm}
        \begin{center}
            \begin{tabular}{c|cc|cc}
            \toprule
            learnable & \multicolumn{2}{c|}{S4}  & \multicolumn{2}{c}{MS3}   \\
            queries & mIoU & F & mIoU & F\\
            \midrule
            \rowcolor{gray!20}
            \checkmark  & 83.1 & 90.5 & 61.3 & 73.0 \\
            $\times$    & 82.7 & 89.9 & 58.5 & 70.9 \\
            \bottomrule
            \end{tabular}
        \end{center}}
    \end{minipage}
}
\hspace{1em}
\subfloat[
    Effect of audio-visual mixer. It is shown that the channel-attention (CHA) mixer works better.
    \label{tab:abla3}
]{
    \centering
    \setlength{\tabcolsep}{0.4mm}
    \begin{minipage}[t]{0.21\linewidth}{
    \vspace{0pt}
      \begin{tabular}{c|cc|cc}
        \toprule
         \multirow{2}{*}{mixer}  & \multicolumn{2}{c|}{S4}  & \multicolumn{2}{c}{MS3}   \\
         & mIoU & F & mIoU & F\\
      \midrule
         $-$  & 81.4 & 88.0 & 59.4 & 70.9 \\
         CRA & 82.7 & 89.7 & 59.8 & 72.2 \\
         \rowcolor{gray!20}
         CHA & 83.1 & 90.5 & 61.3 & 73.0 \\
      \bottomrule
    \end{tabular}}
    \end{minipage}
}
\hspace{1em}
\subfloat[
    Effect of mixing loss. It shows that using mixing loss can indeed improve performance.
    \label{tab:abla4}
]{
    \centering
    \begin{minipage}[t]{0.21\linewidth}{
    \vspace{0pt}
    \setlength{\tabcolsep}{0.2mm}
        \begin{center}
            \begin{tabular}{c|cc|cc}
                \toprule
                 mix  & \multicolumn{2}{c|}{S4}  & \multicolumn{2}{c}{MS3}   \\
                 loss  & mIoU & F  & mIoU  & F    \\
                 \midrule
                 \rowcolor{gray!20}
                 \checkmark  &  83.1   &  90.5  & 61.3 & 73.0 \\
                 $\times$ & 81.3 & 89.1 & 59.6 & 71.3 \\
              \bottomrule
            \end{tabular}
        \end{center}}
    \end{minipage}
}
\caption{AVSegFormer ablation experiments on the S4 and MS3 subsets. 
We report the performance of F-score (denoted as F) and mIoU.
If not specified, the default settings are: 
the number of queries $N_{\rm query}$ is 300, the queries in the decoder are learnable, the audio-visual mixer is used, and the mixing loss is applied. Default settings are marked in \colorbox{baselinecolor}{gray}.}
\label{tab:ablations}
\end{table*}

\begin{figure}[!t]
  \vspace{0.5em}
  \centering
  \includegraphics[width=\linewidth]{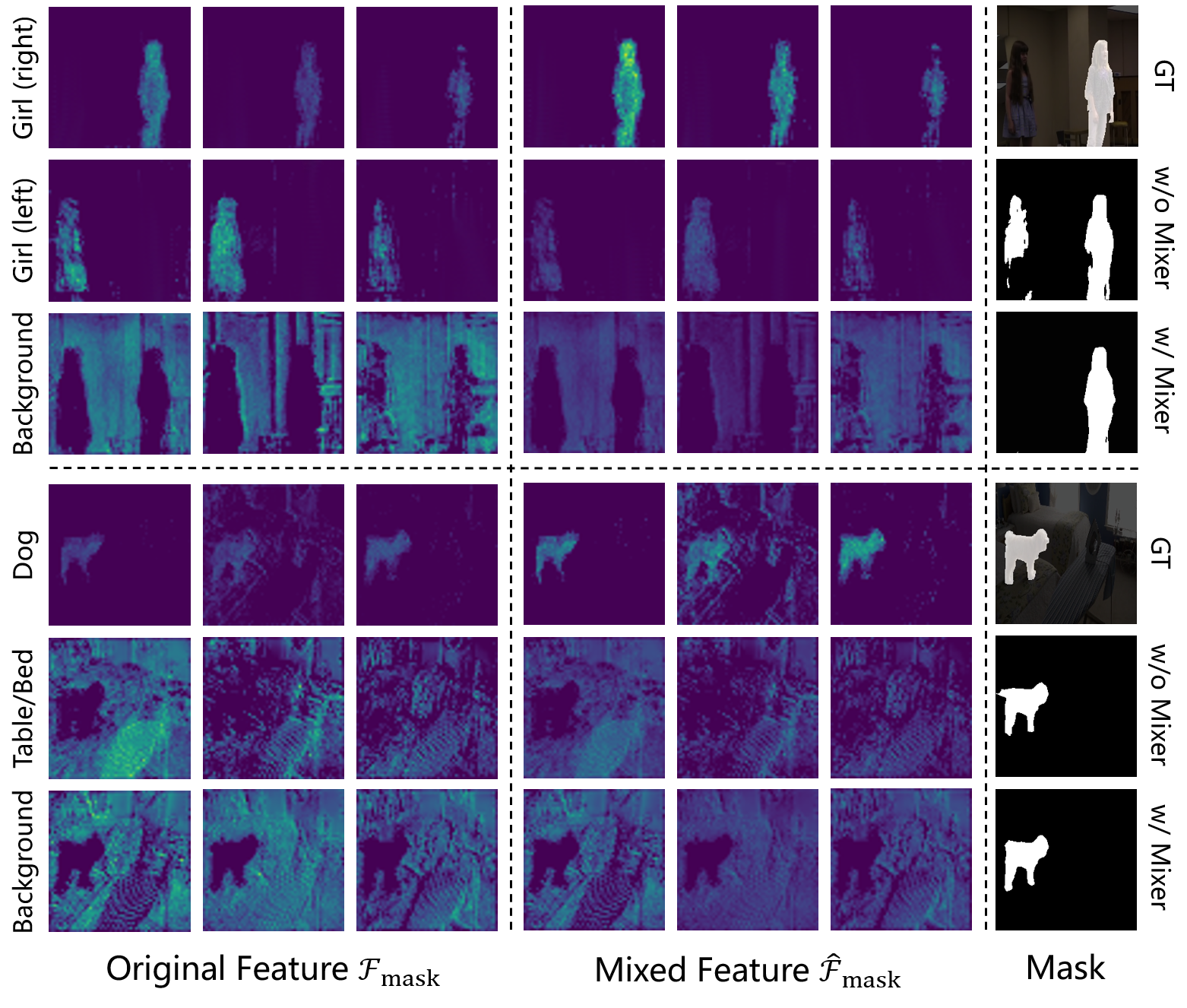}
  \caption{Comparison between the original features $\mathcal{F}_{\rm mask}$ and the mixed features $\mathcal{\hat{F}}_{\rm mask}$. We show two examples with 9 channels with the ground truth and predicted masks. As shown, the features of the ground truth (right girl and dog) are amplified while those of the non-sounding objects (left girl, table/bed, or background) are suppressed, which leads to different segmentation masks. 
  }
  \label{fig:feat}
  \vspace{0.5em}
\end{figure}

\subsection{Dataset}

\textbf{AVSBench-Object \cite{zhou2022audio}} is an audio-visual dataset specifically designed for the audio-visual segmentation task, containing pixel-level annotations.
The videos are downloaded from YouTube and cropped to 5 seconds, with one frame per second extracted for segmentation. 
The dataset includes two subsets: a single sound source subset for single sound source segmentation (S4), and a multi-source subset for multiple sound source segmentation (MS3).
\textbf{S4 subset:} 
The S4 subset contains $4,932$ videos, with $3,452$ videos for training, $740$ for validation, and $740$ for testing. 
The target objects cover 23 categories, including humans, animals, vehicles, and musical instruments. 
\textbf{MS3 subset:}
The MS3 subset includes 424 videos, with 286 training, 64 validation, and 64 testing videos, covering the same categories as the S4 subset.

\textbf{AVSBench-Semantic \cite{zhou2023audio}} is an extension of the AVSBench-Object, which offers additional semantic labels that are not available in the original AVSBench-Object dataset. It is designed for audio-visual semantic segmentation (AVSS).
In addition, the videos in AVSBench-Semantic are longer, with a duration of 10 seconds, and 10 frames are extracted from each video for prediction.
Overall, the AVSBench-Semantic dataset has increased in size by approximately three times compared to the original AVSBench-Object dataset, with 8,498 training, 1,304 validation, and 1,554 test videos.

\begin{figure*}[ht]
  \centering
  \includegraphics[width=\linewidth]{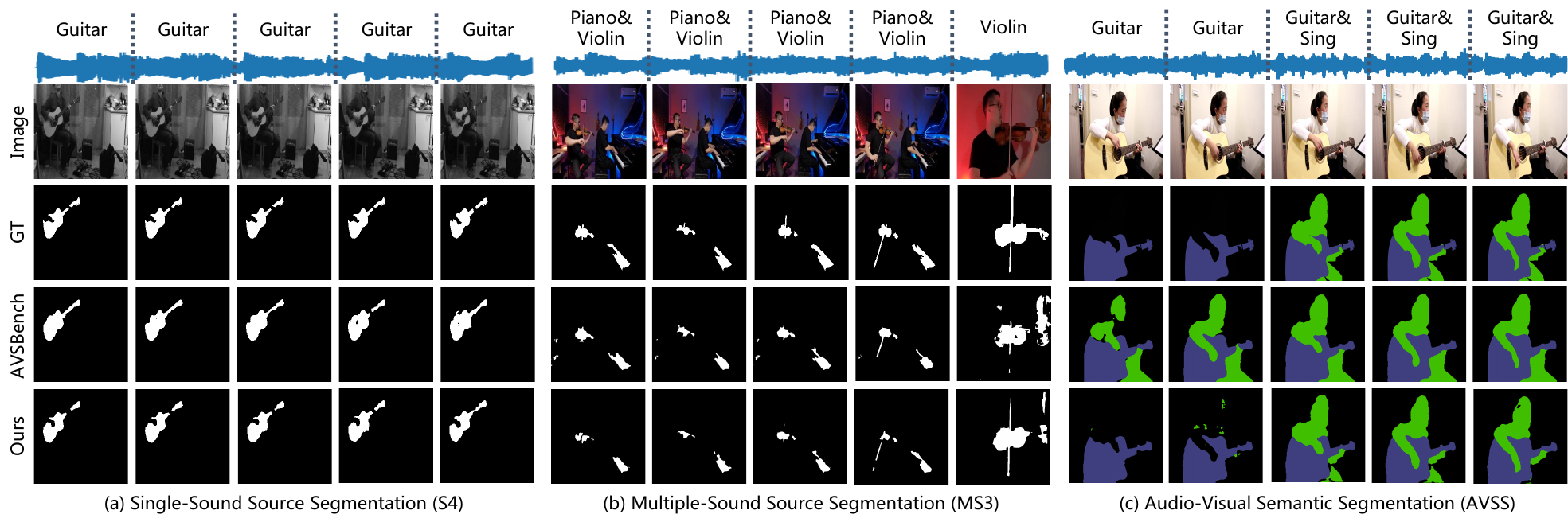}
  \caption{Qualitative results of AVSBench and AVSegFormer on three AVS sub-tasks. These results show that the proposed AVSegFormer can accurately segment the pixels of sounding objects and outline their shapes better than AVSBench.}
  \label{fig:visual}
  \vspace{-0.5em}
\end{figure*}

\subsection{Implementation Details}
\label{sec:detail}
We train our AVSegFormer models for the three AVS sub-tasks using an NVIDIA V100 GPU.
Consistent with previous works \cite{zhou2022audio,zhou2023audio}, we employ AdamW \cite{loshchilov2017decoupled} as the optimizer, with a batch size of 2 and an initial learning rate of $2\times 10^{-5}$.
Since the MS3 subset is quite small, we train it for 60 epochs, while the S4 and AVSS subsets are trained for 30 epochs.
The encoder and decoder in our AVSegFormer comprise 6 layers with an embedding dimension of 256.
We set the coefficient of the proposed mixing loss $\mathcal{L}_{\rm mix}$ to 0.1 for the best performance.
More detailed training settings can be found in the supplementary materials.

\subsection{Comparison with Prior Arts}
We conducted a comprehensive comparison between our AVSegFormer and existing methods \cite{zhou2022audio, zhou2023audio, mao2023contrastive} on the AVS benchmark. 
For fairness, we employ the ImageNet-1K \cite{deng2009imagenet} pre-trained ResNet-50 \cite{he2016deep} or PVTv2 \cite{wang2022pvt} as the backbone to extract visual features, and the AudioSet \cite{gemmeke2017audio} pre-trained VGGish \cite{hershey2017cnn} to extract audio features.

\textbf{Comparison with methods from related tasks.} Firstly, we compare our AVSegFormer with state-of-the-art methods from three AVS-related tasks, including sound source localization (LVS \cite{chen2021localizing} and MSSL \cite{qian2020multiple}), video object segmentation (3DC \cite{mahadevan2020making}, SST \cite{duke2021sstvos} and AOT \cite{yang2021associating}), and salient object detection (iGAN \cite{mao2021transformer} and LGVT \cite{jing_ebm_sod21}).
These results are collected from the AVS benchmark \cite{zhou2022audio}, which are transferred from the original tasks to the AVS tasks. 

As shown in Table~\ref{tab:comp}, our AVSegFormer exceeds these methods by large margins. 
For instance, on the S4 subset, AVSegFormer-R50 achieves an impressive mIoU of 76.45, which is 1.51 points higher than the best LGVT. 
Although LGVT has a better Swin-T \cite{liu2021swin} backbone, our AVSegFormer with ResNet-50 backbone still performs better regarding mIoU. 
In addition, AVSegFormer-PVTv2 produces an outstanding mIoU of 82.06 and an F-score of 89.9 on this subset, which is 7.12 mIoU and 2.6 F-score higher than LGVT, respectively.

On the MS3 subset, AVSegFormer-R50 outperforms the best iGAN with 6.64 mIoU and 8.4 F-score, while AVSegFormer-PVTv2 further raised the bar with an exceptional improvement of 15.47 mIoU and 14.9 F-score. 
On the AVSS subset, our AVSegFormer-R50 yields 24.93 mIoU and 29.3 F-score, and AVSegFormer-PVTv2 obtains an impressive performance of 36.66 mIoU and 42.0 F-score, surpassing AOT by 11.26 mIoU and 11.0 F-score, respectively.

\textbf{Comparison with AVSBench.}
Then, we compare our AVSegFormer with the AVSBench baseline, which is the current state-of-the-art method for audio-visual segmentation.
As reported in Table~\ref{tab:comp}, on the S4 subset, AVSegFormer-R50 achieves 3.66 mIoU and 1.1 F-score improvements over AVSBench-R50, while AVSegFormer-PVTv2 surpasses AVSBench-PVTv2 by 3.32 mIoU and 2.0 F-score.
On the MS3 subset, AVSegFormer-PVTv2 surpasses AVSBench-PVTv2 with a margin of 1.65 mIoU and 5.0 F-score. 
On the AVSS subset, AVSegFormer-R50 and AVSegFormer-PVTv2 achieve significant results with an mIoU improvement of 4.75 and 6.89, and a substantial F-score improvement of 4.1 and 6.8.
These results demonstrate that AVSegFormer outperforms the AVSBench baseline on all sub-tasks, becoming a new state-of-the-art method for audio-visual segmentation.

\subsection{Ablation Study}
In this section, we conduct ablation experiments to verify the effectiveness of each key design in the proposed AVSegFormer. 
Specifically, we adopt PVTv2 \cite{wang2022pvt} as the backbone and conduct extensive experiments on the S4 and MS3 sub-tasks.

\textbf{Number of queries.} To analyze the impact of the number of queries on the model's performance, we conducted experiments with varying numbers of queries for the decoder input, specifically 1, 100, 200, and 300. Our results reveal a positive correlation between the number of queries and the model performance, with the optimal performance obtained when the number of queries was set to 300. Table~\ref{tab:abla1} presents these findings.

\textbf{Effect of learnable queries.}
We further investigated the impact of learnable queries in the decoder inputs. 
As shown in Table~\ref{tab:abla2}, the improvement due to the learnable queries is relatively small in the single sound source task (S4), while it brings significant improvement in the multiple sound source task (MS3). 
This can be attributed to the complexity of sounding objects. We involve a more detailed discussion in the supplementary materials.

\textbf{Effect of audio-visual mixer.}
We then studied the impact of the audio-visual mixer on our model. 
Two versions are designed for this module, as illustrated in Figure~\ref{fig:fusion}. 
The cross-attention mixer (CRA) utilizes visual features as queries and audio features as keys/values for cross-attention, and the channel-attention mixer (CHA) introduced the mechanism of channel attention with audio features as queries and visual features as keys/values. 
As presented in Table~\ref{tab:abla3}, the design of CHA brought greater performance improvement compared to CRA. 

In addition, we also visualize the mask feature before and after the audio-visual mixer, as shown in Figure~\ref{fig:feat}. It is evident that for the sounding object (right girl and dog), the mixer effectively enhanced its features. 
Meanwhile, the non-sounding objects (left girl, table/bed, or background) experienced some degree of suppression. 
These findings align with our hypothesis and further substantiate the effectiveness of the audio-visual mixer.

\textbf{Effect of auxiliary mixing loss.} We finally conducted experiments to learn the impact of the auxiliary mixing loss. 
We train our model with and without the mixing loss, respectively, and report the testing results in Table~\ref{tab:abla4}. It is demonstrated that the auxiliary mixing loss can help a lot in the final prediction.

\textbf{Qualitative analysis.}
We also present the visualization results of AVSegFormer compared with those of AVSBench on three audio-visual segmentation tasks in Figure~\ref{fig:visual}. 
The visualization results clearly demonstrate that our method performs better. It has a strong ability in target localization and semantic understanding and can effectively identify the correct sound source and accurately segment the target object in multiple sound source scenes. 
These results highlight the effectiveness and robustness of our method.

\section{Conclusion}

In this paper, we propose AVSegFormer, a novel audio-visual segmentation framework that leverages the power of transformer architecture to achieve leading performance. 
Specifically, our method comprises four key components: (1) a transformer encoder building the mask feature, (2) a query generator providing audio-conditioned queries, (3) a dense audio-visual mixer dynamically adjusting interested visual features, and (4) a sparse audio-visual decoder separating audio sources and matching optimal visual features.
These components provide a more robust audio-visual cross-modal representation, improving the AVS performance in different scenarios.
Extensive experimental results demonstrate the superior performance of AVSegFormer compared to existing state-of-the-art methods.

\section*{Acknowledgement}
This work is supported by the Natural Science Foundation of China under Grant 62372223.
\bibliography{ref}

\end{document}